\documentclass{article}
\usepackage{spconf,amsmath,amssymb}
\usepackage{graphicx}
\usepackage{booktabs}
\usepackage{algorithm}
\usepackage{algorithmic}
\usepackage{siunitx}
\usepackage{caption}
\usepackage{subcaption}
\usepackage[table]{xcolor}
\usepackage{pgf} 
\usepackage{booktabs}
\usepackage{array}
\usepackage{url}

\newcommand{\swatchOD}[3]{%
  \pgfmathsetmacro{\rr}{exp(-#1)}%
  \pgfmathsetmacro{\gg}{exp(-#2)}%
  \pgfmathsetmacro{\bb}{exp(-#3)}%
  \begingroup
  \setlength{\fboxsep}{5.7pt}%
  \raisebox{.6ex}{\fcolorbox{black!30}[rgb]{\rr,\gg,\bb}{\rule{2.8ex}{-1.2ex}}}%
  \endgroup
}

\newcommand{\rgbtriplet}[3]{\(\,#1,\,#2,\,#3\,\)}

\title{Beer-Lambert Autoencoder for Unsupervised Stain Representation Learning and Deconvolution in Multi-immunohistochemical Brightfield Histology Images}

\name{Mark Eastwood$^1$ \qquad Thomas McKee$^2$ \qquad Zedong Hu$^3$ \qquad Sabine Tejpar$^3$ \qquad Fayyaz Minhas$^1$ \qquad}
\address{$^1$University of Warwick \\
$^2$University of Geneva \\
$^3$Katholeike Universitat Leuven}

\begin{document}
\maketitle

\begin{abstract}
Separating the contributions of individual chromogenic stains in RGB histology whole slide images (WSIs) is essential for stain normalization, quantitative assessment of marker expression, and cell-level readouts in immunohistochemistry (IHC). Classical Beer-Lambert (BL) color deconvolution is well-established for two- or three-stain settings, but becomes under-determined and unstable for multiplex IHC (mIHC) with $K>3$ chromogens. We present a simple, data-driven encoder-decoder architecture that learns cohort-specific stain characteristics for mIHC RGB WSIs and yields crisp, well-separated per-stain concentration maps. The encoder is a compact U-Net that predicts $K$ nonnegative concentration channels; the decoder is a differentiable BL forward model with a \emph{learnable} stain matrix initialized from typical chromogen hues. Training is unsupervised with a perceptual reconstruction objective augmented by loss terms that discourage unnecessary stain mixing. On a colorectal mIHC panel comprising 5 stains (H, CDX2, MUC2, MUC5, CD8) we show excellent RGB reconstruction, and significantly reduced inter-channel bleed-through compared with matrix-based deconvolution. Code and model are available at \url{https://github.com/measty/StainQuant.git}.
\end{abstract}

\section{Introduction}
Separating histological stains from brightfield RGB is a long-standing problem. Under the BL law, absorbances are additive in optical density (OD), enabling linear unmixing given a stain matrix whose columns are stain OD vectors \cite{ruifrok2001quantification}. Practical variants estimate the matrix from data (e.g.\ by singular vector directions \cite{macenko2009method} or sparse nonnegative matrix factorization with structure preservation \cite{vahadane2016structure}). These approaches work well for H\&E or H\&E{+}DAB, where $K\leq3$, but multiplex IHC frequently uses $K>3$ chromogens whose spectra overlap in RGB \cite{tan2020overview,bosisio2022nextgen}. Then, pixel-wise linear unmixing is under-determined, yielding poor separation and significant cross-over between channels. Utilizing stronger priors based on domain knowledge can help address this.

For separation in brightfield IHC imaging, existing methods include optimization with group sparsity for brightfield mIHC \cite{chen2015group}, weakly-supervised CNNs trained from dot annotations \cite{abousamra2020weakly}, Bayesian/DIP-style blind deconvolution for conventional stains \cite{yang2023deep}, physics-guided zero-shot deconvolution and normalization \cite{chen2025pgdips}, and an unsupervised inversion-regularized formulation using color samples \cite{abousamra2024unsupervised}. 

Most existing approaches rely on some form of supervision (e.g.\ dot annotations), assume fixed stain matrices, or do not learn morphological cues. The method most similar to ours is the unsupervised inversion-regularized autoencoder for mIHC \cite{abousamra2024unsupervised}, which also performs reconstruction under the BL model and can exploit morphological context, but this keeps the stain matrix fixed and constrains separation via an inverse-image penalty. In contrast the stain basis is a parameter of our model that we \emph{learn} jointly with concentrations under a soft color prior. We pair this with a richer, unsupervised loss formulation that encourages sharper, low-overlap stain maps and higher-fidelity reconstructions. Unfortunately code for comparative evaluation could not be found in the github associated with the paper.

Our main contribution is a compact encoder-decoder stain-separation model whose design goals are: (1) a physically grounded decoder, and (2) clean separation promoting low-overlap channels wherever this is consistent with the output RGB. Our model is trained using an unsupervised objective incorporating priors based on domain knowledge, and generates $K$ well-separated stain channels from an input RGB mIHC. We present qualitative and quantitative results on an mIHC panel for colorectal cancer including H, CDX2, MUC2, MUC5, and CD8, showing that separation improves on classical matrix-based deconvolution.

Multiplex IHC staining provides a cost-effective, simpler alternative to multiplex immunofluorescence while still capturing rich biological detail. However, quantitative analysis remains difficult due to stain cross-talk in RGB brightfield images. Our approach, which achieves sharp and consistent stain separation, may enable broader and more quantitative use of mIHC protocols in both research and clinical settings.

\begin{figure*}[]
\centering
\includegraphics[width=0.86 \textwidth]{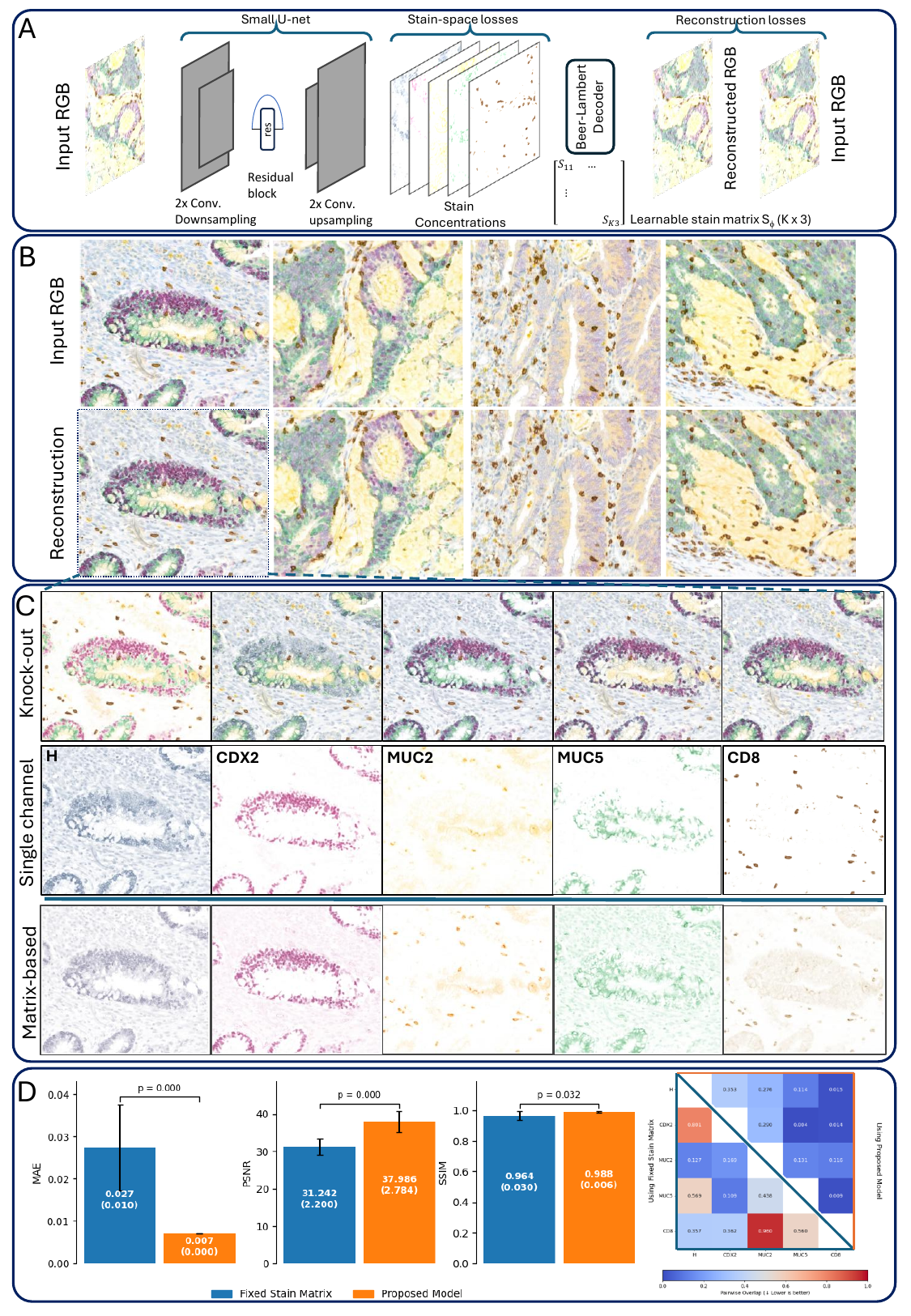}
\caption{\textbf{Method and results.} \emph{A:} Encoder-decoder with learnable stain matrix. \emph{B:} Examples of Input patches and model reconstructions from channel decomposition. \emph{C:} Single-channel per-stain concentration map (H, CDX2, MUC2, MUC5, CD8) renderings and knock-out (KO) images, and comparison with single channel renders from matrix-based deconvolution (bottom row). \emph{D:} Reconstruction metrics and pairwise channel similarity comparison.}
\label{fig:qual}
\end{figure*}

\section{Methods}
\paragraph*{Stain Separation Model.}
Let $x\in[0,1]^3$ denote a normalized RGB pixel and $o=-\log(x+\epsilon)$ its channel-wise OD. For $K$ chromogens, let $S\in\mathbb{R}_{+}^{3\times K}$ be the stain OD matrix with columns $s_k$, and $c\in\mathbb{R}_{+}^K$ the corresponding per-pixel nonnegative stain concentrations. The BL forward model \cite{ruifrok2001quantification} is
\begin{equation}
\hat{o} = Sc,\qquad \hat{x}=\exp(-\hat{o}). \label{eq:forward}
\end{equation}
When $K>3$, the Beer–Lambert system is underdetermined; even if the stain matrix $S$ were known, a single RGB pixel does not uniquely specify the stain concentrations $c$. However, at the \emph{dataset} level this improves with priors: (i) $S$ is shared across the cohort; (ii) $c$ is usually sparse and low-entropy per pixel; (iii) morphology provides cues to expected stain affinities. Our learning objective and architecture exploit these priors.

\paragraph*{Architecture.}
We use a small U-Net encoder $E_\theta$ with two downsampling convolution blocks, a residual bottleneck, and two upsampling conv blocks with skip connections (see Fig. \ref{fig:qual} A) that maps an RGB patch $X\in[0,1]^{H\times W\times 3}$ to concentrations $C=E_\theta(X)\in\mathbb{R}_{+}^{H\times W\times K}$. Non-negativity is enforced by a final $\mathrm{softplus}$. The decoder implements \eqref{eq:forward} with a learnable $S_\phi\in\mathbb{R}_{+}^{3\times K}$.
$S_\phi$ is initialized by $\tilde{S}$, unit-normalized OD vectors corresponding to typical chromogen colors, which we estimated manually by looking at pixel values of small regions that look to be 'pure stains' in QuPath \cite{qupath2017}.

\subsection{Objective and loss terms}
The objective of the model is to separate input RGB images into a sparse $K$ channel concentration map $C$ which yields a faithful reconstruction $\hat{X}$ of the input image when passed through the physics-grounded BL decoder. Training minimises a sum of five terms:
\[
\mathcal{L}
= \mathcal{L}_{\mathrm{rec}}
+\lambda_{\mathrm{ent}}\mathcal{L}_{\mathrm{ent}}
+\lambda_{\mathrm{col}}\mathcal{L}_{\mathrm{col}}
+\lambda_{\mathrm{ov}}\mathcal{L}_{\mathrm{ov}}
+\lambda_{\mathrm{mask}}\mathcal{L}_{\mathrm{mask}}
\]
described in more detail below.
 
\paragraph*{Reconstruction loss.}
A composite reconstruction term enforces fidelity in the image domain and in a perceptual feature space:
\begin{equation}
\mathcal{L}_{\mathrm{rec}}
= \left\lVert \hat{\mathbf{X}} - \mathbf{X} \right\rVert_{1}
+ 2\cdot
\frac{1}{L}\sum_{\ell=1}^{L}
\big\lVert \phi_\ell(\hat{\mathbf{X}}) - \phi_\ell(\mathbf{X}) \big\rVert_{1}.
\label{eq:lrec}
\end{equation}
Here $\phi_\ell(\cdot)$ are feature maps from a fixed VGG-19 network at layers $2, 7, 16, 27$, and $L$ is the number of tapped layers; the second term is the average $\ell_1$ distance across these layers.

\paragraph*{Activation-entropy loss.}
To discourage stain channel crossover that is not necessary to reconstruct input pixels, we minimise the Shannon entropy of the per-pixel, channel-normalised concentration distribution, while ignoring background/near-zero pixels. Define
\[
s(x)=\sum_{k=1}^{K} C_k(x),\qquad
p_k(x)=\frac{C_k(x)}{s(x)+\epsilon}
\]
Then
\begin{equation}
\mathcal{L}_{\mathrm{ent}}
=\frac{1}{|\Omega_\tau|}
\sum_{x\in\Omega_\tau}
\Big(-\sum_{k=1}^{K} p_k(x) \log\big(p_k(x)+\epsilon\big)\Big).
\label{eq:lent}
\end{equation}
where $\Omega_\tau=\{x:\ s(x)>\tau\}$ is the set of non-background pixels (estimated by a simple threshold $\tau=10^{-2}$).

\paragraph*{Colour-consistency loss.}

The learnable basis $S_\phi$ in the decoder is softly anchored to the initial estimate $\tilde{S}$ via an elementwise $\ell_1$ mismatch:
\begin{equation}
\mathcal{L}_{\mathrm{col}}
=\frac{1}{3K} \lVert S_\phi - \widetilde{S}\rVert_{1}
\label{eq:lcol}
\end{equation}
This ensures the stain channels retain their identity, and incorporates prior knowledge of typical chromogen hues.

\paragraph*{Top-$k$ overlap loss.}
Let $p\in(0,1)$ be a small fraction (default $p=0.05$). For each channel $k$, form the set $S_k$ of the top-$p$ fraction of pixels by $C_k(x)$. Define the per-pixel multi-channel overlap count
$o(x)=\max\big(0,\sum_{k=1}^{K}\mathbf{1}_{x\in S_k}-1\big)$.
The penalty is
\begin{equation}
\mathcal{L}_{\mathrm{ov}}
= \frac{1}{p|X|} \sum_{x\in X} o(x),
\label{eq:lov}
\end{equation}
which discourages pixels from being among the strongest activations of multiple stains; the $1/p$ factor approximately normalises the scale across choices of $p$.

\paragraph*{Mask-dominance loss.}
For stains that produce sparse, high-intensity signals, a crude heuristic binary mask $M\in\{0,1\}^{H\times W}$ can be constructed on the fly (via hue thresholding based on initial hue estimate) to flag candidate positive pixels, and a channel index $c^\star\in\{1,\dots,K\}$ designated as the corresponding stain. Then, the following loss encourages the designated channel to dominate masked pixels:
\begin{equation}
\mathcal{L}_{\mathrm{mask}}
=\frac{1}{|\Omega_M|}\sum_{x\in\Omega_M}
\Big(1 - \frac{C_{c^\star}(x)}{\sum_{k=1}^{K} C_k(x) + \epsilon}\Big),
\label{eq:lmask}
\end{equation}
where $\Omega_M$ is the set of pixel locations for which the mask is $1$. In the case of our mIHC panel, we apply this only to the CD8 stain, which stains only certain relatively rare immune cells but is very intense and therefore amenable to thresholding.

\begin{table}[t]
\centering
\small
\setlength{\tabcolsep}{2pt} 
\renewcommand{\arraystretch}{1.1}
\begin{tabular}{
  l
  @{\hspace{1ex}} c @{\hspace{0.5ex}} c
  @{\hspace{1.5ex}} c @{\hspace{0.5ex}} c
}
\toprule
 & \multicolumn{2}{c}{\textbf{Initial}} &
   \multicolumn{2}{c}{\textbf{Learned}} \\
\textbf{Stain} & \textbf{(R,G,B)} & \textbf{Color} &
                 \textbf{(R,G,B)} & \textbf{Color} \\
\midrule
H    & \rgbtriplet{0.620}{0.637}{0.458} &
         \swatchOD{0.620}{0.637}{0.458} &
       \rgbtriplet{0.705}{0.581}{0.408} &
         \swatchOD{0.705}{0.581}{0.408} \\
CDX2 & \rgbtriplet{0.290}{0.832}{0.473} &
         \swatchOD{0.290}{0.832}{0.473} &
       \rgbtriplet{0.242}{0.843}{0.480} &
         \swatchOD{0.242}{0.843}{0.480} \\
MUC2 & \rgbtriplet{0.033}{0.343}{0.939} &
         \swatchOD{0.033}{0.343}{0.939} &
       \rgbtriplet{0.028}{0.239}{0.971} &
         \swatchOD{0.028}{0.239}{0.971} \\
MUC5 & \rgbtriplet{0.741}{0.294}{0.604} &
         \swatchOD{0.741}{0.294}{0.604} &
       \rgbtriplet{0.737}{0.300}{0.606} &
         \swatchOD{0.737}{0.300}{0.606} \\
CD8  & \rgbtriplet{0.300}{0.491}{0.818} &
         \swatchOD{0.300}{0.491}{0.818} &
       \rgbtriplet{0.330}{0.536}{0.777} &
         \swatchOD{0.330}{0.536}{0.777} \\
\bottomrule
\end{tabular}
\caption{Initial and learned stain basis OD vectors, with color patches
showing the corresponding transmitted RGB.}
\label{tab:stain_matrices}
\end{table}

\section{Experiments and Results}

We train on 32,547 non-overlapping 960$\times$960 patches at \SI{0.5}{\micro\meter/px} extracted from colorectal cancer WSIs stained with H, CDX2, MUC2, MUC5, CD8. Patches were extracted from within the tissue-mask (otsu luminosity threshold) over $128$ WSIs. The mIHC panel used has been chosen to provide complementary information on tumour differentiation and immune context. Hematoxylin (H) provides nuclear morphology and overall tissue architecture; CDX2 marks intestinal epithelial lineage commitment and is frequently reduced or lost in poorly differentiated carcinomas; MUC2 and MUC5 identify secretory goblet-cell and mucinous phenotypes, respectively, which can signal reversion toward fetal-like or mucin-rich differentiation programs; and CD8 highlights cytotoxic T-cell infiltration, reflecting the immune microenvironment and anti-tumour response. Together, these markers expose valuable insights into tumor biology.

\paragraph*{Stain channel crossover.}

To quantify the level of separation of the stains, we calculate the per-image cosine similarity between each pair of stain channels. The mean over a sample of images is shown in Fig. \ref{fig:qual} panel D. We can see that matrix-based deconvolution shows significant crossover between channels; particularly egregious is the large bleed-through between H and MUC5, and CD8 and MUC2, neither of which we would expect as they stain very different tissue. Our model displays more cleanly separated stains, with the highest co-occurrence between H and CDX2, which is expected as both are nuclear stains that will co-stain some epithelial nuclei.

\paragraph*{The learned stain vectors.}
In table \ref{tab:stain_matrices} we can see the initial and final stain vectors that comprise the initial stain matrix $\tilde{S}$ and the matrix learned by the model during training $S_\phi$. As we can see, while each stain has retained its identity, adjustments have been made to each with the largest being made to MUC2 and H. These stains also showed the most channel channel crossover (H with CDX2 and MUC5, and MUC2 with CD8) in the plot in Fig \ref{fig:qual} panel D, which is consistent with those stain vectors diverging more during training. 

\paragraph*{Reconstruction and visualization.} The model achieves excellent RGB reconstruction, indicating that the learned $(S_\phi,C)$ form a consistent representation of cohort stains. Concentration maps are visually crisp with sharply delineated structures per stain, and reduced bleed-through compared to matrix-based deconvolution (Fig.~\ref{fig:qual}). This is most evident for CD8 and MUC5: the learned CD8 concentrates on lymphocytes without the cross-over with MUC2 seen using matrix-based separation, while in the case of MUC5 the crossover with the H channel is significantly reduced. This is also quantified in Fig. \ref{fig:qual} panel D, showing improvements in all metrics.

Single-channel renderings (zeroing all but one channel before the decoder) produce sensible chromogen images for each stain, facilitating interpretation and downstream quantification. KO images (zeroing one channel) suppress only the expected structures (e.g.\ CD8+ cells), and show minimal collateral changes, supporting low cross-talk. Examples can be seen in Fig. \ref{fig:qual}.

\section{Discussion \& Conclusions}
Pixel-wise unmixing is under-determined, but learning $(S_\phi,E_\theta)$ over a large set of examples enables us to incorporate desirable priors and learn morphological cues to improve stain separation. We introduced a model for mIHC stain separation that scales beyond the BL limit of $K\le 3$, using a physically grounded BL decoder, and a convolutional encoder that contributes spatial context and shape priors through its receptive field. Overlap and entropy penalties act as regularizers encouraging structured sparsity that reduces cross-talk by discouraging unnecessary stain overlap; the primary reconstruction loss ensures we allow stain mixing where it is needed to reconstruct the observed image. On a colorectal panel (H, CDX2, MUC2, MUC5, CD8) it delivers excellent reconstruction and visibly crisper, low-overlap stain maps compared to matrix-based deconvolution. We hope the availability of strong stain separation tools will enable broader and more quantitative use of such mIHC protocols.

The proposed approach has some limitations. The learned $S$ is panel- and cohort-specific, so transfer across labs may require fine-tuning, and further testing over a wider range of mIHC panels is needed to ensure the approach is robust across a wide range of staining protocols. The Hue masks for rare stains are crude and panel-dependent; while they only weakly steer rare channels, errors in mask generation if such stains are not sufficiently prominent may make this loss component less useful. Absolute quantification is not guaranteed by our method, as it does not use (and we do not have) explicit per-stain ground truth; we target \emph{relative} separation. 

The model and code (available at \url{https://github.com/measty/StainQuant.git}) enables labs to train a model providing concentration maps for their own mIHC cohort.

\section*{Compliance with Ethical Standards}
The data used in this study was approved by the ethics committee of Katholeike Universitat Leuven (references S62294 and S64121).

\section*{Acknowledgments}

FM and ME acknowledge funding support from EPSRC Grant EP/W02909X/1. ST acknowledges funding support from a BOF-Fundamental Clinical Research mandate (FKO) from KU Leuven (DOC/CM21-FKO-02) and KULeuven internal research project C24M/22/055. The authors have no other relevant financial or non-financial interests to disclose.

\small
\bibliographystyle{IEEEbib}
\bibliography{refs}

\end{document}